# Reversible Upper Confidence Bound Algorithm to Generate Diverse Optimized Candidates

Bin Chong[1], Yingguang Yang[2], Zi-Le Wang[3], Hang Xing[1], and Zhirong Liu[1,*]

Dedicated to the 100th Birthday of Professor Youqi Tang

[1] College of Chemistry and Molecular Engineering, and Beijing National Laboratory for Molecular Sciences (BNLMS)
Peking University
Beijing 100871 (China)
[2] School of Cyberscience
University of Science and Technology of China
Hefei 230026 (China)
[3] State Key Laboratory of Low Dimensional Quantum Physics, Department of Physics
Tsinghua University
Beijing 100084 (China)

Correspondence and requests for materials should be addressed to Z.L.(e-mail: LiuZhiRong@pku.edu.cn).

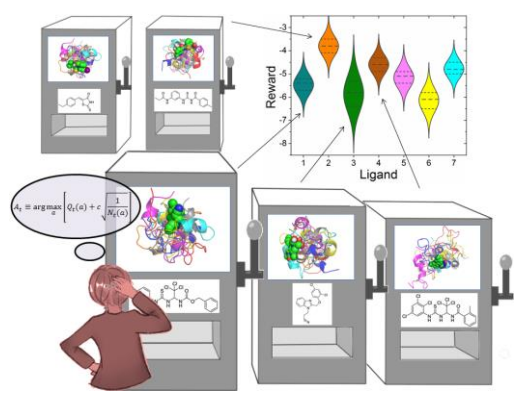

**Abstract:** Most algorithms for the multi-armed bandit problem in reinforcement learning aimed to maximize the expected reward, which are thus useful in searching the optimized candidate with the highest reward (function value) for diverse applications (e.g., AlphaGo). However, in some typical application scenaios such as drug discovery, the aim is to search a diverse set of candidates with high reward. Here we propose a reversible upper confidence bound (rUCB) algorithm for such a purpose, and demonstrate its application in virtual screening upon intrinsically disordered proteins (IDPs). It is shown that rUCB greatly reduces the query times while achieving both high accuracy and low performance loss.The rUCB may have potential application in multipoint optimization and other reinforcement-learning cases.

In despite of the lack of unique structure under physiological conditions, intrinsically disordered proteins (IDPs) exhibit prevalent key roles in diverse biological processes.[1-7] IDPs are widely related to human diseases such as cancer, neurodegeneration and diabetes,[8-11] so they have been long been recognized as attractive therapeutic targets.[12-16] However, drug design upon IDPs is extremely challenging because of their dynamic nature.[17-22] In contrast to usual ordered proteins, IDPs exist in an ensemble of highly heterogeneous conformations which is difficult to be characterized via experimental or computational approaches.[23-26] Small molecules bind to IDPs through dynamic and transient ("fuzzy") interactions,[18,27,28] and thus modulate the disordered conformational ensemble to realize molecular recognition. Even if the conformation ensembles were determined,[18,29-33] it would be computationally prohibitive in computer-aided drug design to dock usual large ligand libraries (with ~$10^5$ or even more compounds) to thousands and tens of thousands of conformations.[34-36]

In recent years, Reinforcement learning has attracted great interest due to its capacity in interactively exploring the unknown state space,[37,38] and some effort has been attempted in the field of chemistry.[39-45] For typical problems of reinforcement learning such as the game of Go, the space of trajectory/state is vast and it is thus infeasible to conduct exhaustive search. However, an effective algorithm need not explore the whole space accurately; it only need focus on the important part, e.g., the moves with high value in Go. Virtual screening of ligands upon IDPs shares a very similar situation: the space of (IDPs conformation, ligand) is vast, but it needs only to pick out a tiny minority of ligands which have high affinity to IDPs. Therefore, reinforcement learning may provide an efficient solution to the bottleneck of virtual screening upon IDPs.

Here, we develop an efficient reinforcement learning approach to accomplish the virtual screening for drug design on IDPs. Using the algorithm, the docking number is greatly reduced without affecting the performance of virtual screening, e.g., the average docking number per ligand can be as small as 2. This suggests virtual screening upon IDPs can be basically solved with a computation cost comparable to that for ordered proteins.

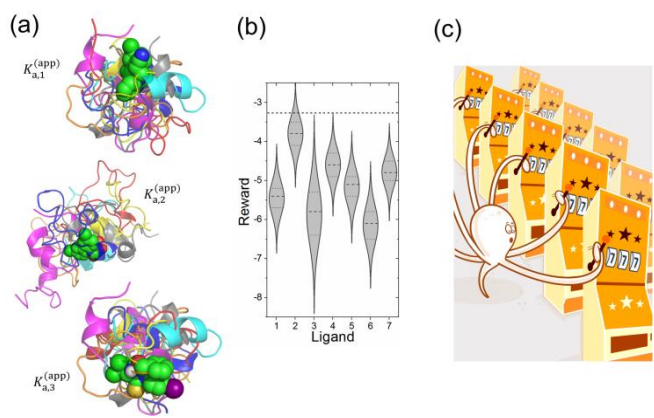

**Figure 1.** Schematic illustration on (a) virtual screening upon IDPs, (c) multi-armed bandit problem, and (b) the (reversible) UCB algorithm connecting them.

The problem of virtual screening upon IDPs is schematically shown in Figure 1(a). The ensemble of an IDP is composed of a series of conformations described by statistical thermodynamics. The small-molecule library contains $N$ different ligands. The binding equilibrium between each conformation-$i$ and ligand-$j$ is described by an association constant being related to the binding free energy $\Delta G_{i,j}$ as

$$K_{a,i,j} = e^{-\frac{\Delta G_{i,j}}{RT}}. \tag{1}$$

$\Delta G_{i,j}$ is estimated from molecular docking as in usual rational drug design.[46-48] Overall, the apparent association constant between the protein and the ligand-$j$ is given by

$$K_{a,j}^{(app)} = \frac{\sum_i e^{-\frac{\Delta G_{i,j}}{RT}}}{\sum_i 1}. \tag{2}$$

The aim of virtual screening is to pick out top $m$ ligands with $K_{a,j}^{(app)}$ larger than those of the others, usually with $m \ll N$. In practice, a ligand is docked to only a finite number of conformations instead of the whole ensemble, which inevitably leads to the uncertainty of estimated $K_{a,j}^{(app)}$ and thus the decrease of accuracy.

The above problem is inherently related to the well-known multi-armed bandit problem in reinforcement learning [Figure 1(c)]: there is a slot machine with $k$ levers (arms), and each play of one of the levers will lead to a reward following a certain probability distribution; the objective of the game is to maximize the expected total reward over some time period. Obviously, the best action is to always select the lever with the highest expected reward. However, the player does not know the expected values in advance, and has to estimate them from the outcome of previous actions. An efficient algorithm for the multi-armed bandit problem is the so-called upper confidence bound (UCB) algorithm,[49,50] where at each round-$t$ the lever-$j$ to be played is chosen to maximize an indexed function as:[37]

$$\arg\max_j \left[Q_j(t) + c\sigma_j(t)\sqrt{\frac{\ln t}{n_j(t)}}\right]. \tag{3}$$

$Q_j(t)$ and $\sigma_j(t)$ are the mean and distribution width of the reward of lever-$j$, respectively, being estimated from the previous results. $n_j(t)$ denotes the number of times that lever-$j$ has been selected prior to round-$t$. The idea is that the second term is a measure of the uncertainty in the estimate of the lever's expected reward, and the function to be maximized is thus a sort of upper bound on the possible true value of the lever-$j$ [see also Figure 1(b)], with the parameter $c$ determining the confidence level. When $t \to +\infty$, it can be proven mathematically[49,50] that $n_j(t) \to +\infty$ for any $j$ and the selection converges to the best action with a probability of 1.

When the number of top ligands to be picked is $m = 1$, the problem of virtual screening upon IDPs is equivalent to the multi-armed bandit problem, and the UCB algorithm in Eq. (3) can be directly adopted. In usual tasks, however, $m$ is not equal to 1, so the algorithm should be properly modified. In addition, the task of virtual screening upon IDPs differs from multi-armed bandit problem at least in two aspects: firstly, the ligand number ($N$) is much larger than the lever number; secondly, the total docking number is not going to infinity, i.e., we are not interested in the ultimate performance under $t \to +\infty$, but focus on improving efficiency in reasonable time, e.g., with $t \leq 2N$. For these sakes, we propose the following reversible UCB (abbreviate as rUCB) algorithm for reinforcement-learning virtual screening upon IDPs (see the Supporting Information for more details):

**Initialization:** Dock each ligand once.
**Loop ($t = N + 1, T$):** Choose one ligand to dock once by
— Pre-choose $m$ ligands that maximize an indexed function as

$$\arg\max_j \left[\log K_{a,j}^{(app)}(t) + c\sigma_j(t)\sqrt{\frac{1}{n_j(t)}}\right], \tag{4}$$

where $Q_j = \log K_{a,j}^{(app)}$ is adopted, and $\ln t$ in Eq. (3) of UCB is discarded to improve efficiency;
— Among $m$ pre-chosen ligands, choose the one that minimize another indexed function as

$$\arg\min_j \left[\log K_{a,j}^{(app)}(t) - c\sigma_j(t)\sqrt{\frac{1}{n_j(t)}}\right], \tag{5}$$

and dock it once.

**Final prediction:** Pick $m$ ligands that maximize $\log K_{a,j}^{(app)}(T)$ as the predicted top ligands from virtual screening.

As will be shown below, the idea of the algorithm is to reduce the uncertainty near the boundary between top-$m$ ligands and the others.

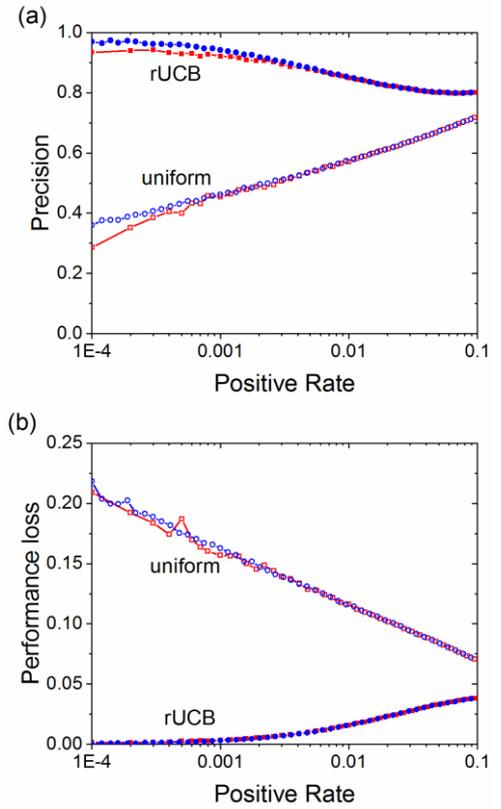

**Figure 2.** Effects of the rUCB algorithm (filled symbols) in comparison with those of the uniform algorithm (opened symbols). Synthetic datasets with $N = 10^4$ (red) and $N = 10^5$ (blue) were used. The average docking times per ligand is 2, i.e., $T/N = 2$. The parameter $c$ in rUCB has been optimized (Figure S1). Each data point was averaged from 200-500 runs of simulation to get better statistics.

We first test the algorithm on synthetic datasets, which were constructed by assuming $\Delta G_{i,j}$ for each ligand $j$ to obey Gaussian distribution with random parameters being inferred from a real dataset-I of complete docking between 283 ligands and 16716 conformations of the oncoprotein c-Myc (see the Supporting Information). We used two metrics to measure the quality of algorithms: the precision as the ratio of the predicted top-$m$ ligands belonging to true top-$m$ ligands, and the performance loss defined as the difference between the average $\log K_{a,j}^{(app)}$ of the true top-$m$ ligands and the average true $\log K_{a,j}^{(app)}$ of the predicted top-$m$ ligands. In addition to the proposed rUCB algorithm, for comparison, we also adopt a trivial (non-learning) uniform algorithm where each ligand is docked with the same number of IDP conformations, which was widely used in previous practice.[34-36]

The reinforcement-learning algorithm has extraordinary effects as demonstrated in Figure 2. With a usual virtual-screening scale of picking out top-100 ligands from a library of $10^5$ ligands (with a positive rate of 0.1%), the precision of rUCB algorithm is as high as 94% when $T = 2N$, i.e., a ligand is docked only two times on average. With the same docking times, the usual uniform algorithm can only achieve a precision of 47%. The improvement caused by reinforcement learning is even more pronounced when one examines the performance loss [Figure 2(b)]: it decreases from 0.16 of the uniform algorithm dramatically to 0.003 of rUCB algorithm. This means that even though rUCB does not achieve the precision limit of 100%, the performance loss is nearly negligible in this case. When the designated positive rate gets higher, the quality of rUCB decreases while that of the uniform algorithm increases. However, the superiority of rUCB remains, e.g., under a positive rate of 1%, the precision is 85% *vs.* 57% and the performance loss is 0.0095 *vs.* 0.12. To achieve performances similar to those of rUCB under $T/N = 2$, the uniform algorithm has to dock each ligand 30 (for a positive rate of 1%) or 200 (for a positive rate of 0.1%) times (Figure S2). The performances increase slightly with increasing the ligand number $N$ under the same positive rate (Figure S3). These trends are favorable for actual applications of rUCB since virtual screening is usually conducted with large ligand library and small positive rate.

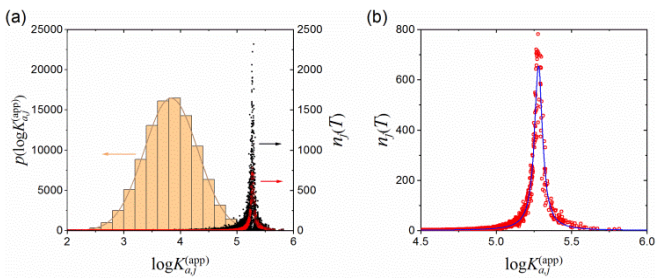

**Figure 3.** Distribution of docking times under rUCB. Synthetic dataset with $N = 10^5$ was used. $m = 100$ to give a positive rate of 0.1%. $T/N = 2$. (a) Histogram of $\log K_{a,j}^{(\text{app})}$ (histogram) and the simulated docking times $n_j(T)$ of ligands (scattering symbols, where the black come directly from 10 simulations while the red is the average from 400 simulations). (b) The docking times $n_j(T)$ averaged from 400 simulations (red circles), where the blue line is the fitted Cauchy-Lorentz function.

To understand the reason why rUCB is so efficient in virtual screening, we analyse the distribution of docking times $n_j(T)$ among ligands (Figure 3). $n_j(T)$ as a function of $\log K_{a,j}^{(\text{app})}$ has a sharp peak, the position of which drastically deviates from the histogram peak of $\log K_{a,j}^{(\text{app})}$ [Figure 3(a)]. The majority of ligands have very few docking times, but some are docked by hundreds to thousands of times. This reflects the central idea for efficient virtual screening algorithms on IDPs: to use the estimated $\log K_{a,j}^{(\text{app})}$ to separate top-$m$ ligands from the others, $\log K_{a,j}^{(\text{app})}$ of ligands closer to the boundary should be determined more accurately. Therefore, for ligands far away from the boundary, very few times (one or two times) of docking are sufficient to classify them with high confidence; while for those close to the boundary, more docking times are required. The docking times should increase with decreasing the distance to the boundary. Although large fluctuations exist among different runs of simulations [black scatterings in Figure 3(a)], the average behavior [black scatterings in Figure 3(a,b)] matches this central idea very well. Detailed analysis (see the Supporting Information) suggests that $n_j$ in rUCB obeys the Cauchy-Lorentz function, which is demonstrated in Figure 3(b) as the fitted curve (see also Figure S4 for results under other positive rates).

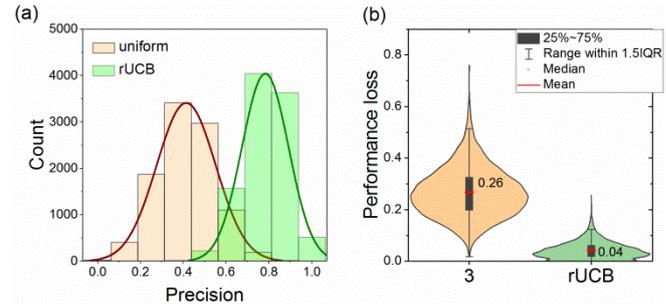

**Figure 4.** Validation of rUCB on a real dataset with 828 ligands. $m = 8$ to give a positive rate of 1% roughly. $T/N = 2$. 10000 runs of simulation with different random seeds were conducted for statistics. (a) Histogram of the precision of rUCB and the uniform algorithm. (b) Distribution of the performance loss in violin plot with box chart.

To validate the effect of rUCB on real dataset, we constructed a real dataset-II with 828 ligands completely docked with 16716 conformations of c-Myc (see the Supporting Information), and test the effects of rUCB and the uniform algorithm under a positive rate of 1% (Figure 4). The obtained precision of rUCB is 78% in average, which is much higher that of the uniform algorithm (41%). Their distribution widths are comparable [Figure 4(a)]. The overwhelming advantage of rUCB is more obvious in terms of the performance loss: 0.04 vs. 0.27 [Figure 4(b)]. With decreasing the positive rate, the advantage of rUCB would be further enhanced (Figure S5). In addition, the docking times of rUCB can be reduced, e.g., to $T/N = 1.5$ to yield a precision of 67% and a performance loss of 0.09 (Figure S5). It is noted that the performance loss in real dataset is larger than that in synthetic dataset no matter which algorithm (rUCB or the uniform) is used, which may due to the assumption of Gaussian distribution in synthetic dataset. However, the advantage of rUCB keeps robust.

Application of rUCB was demonstrated on a large library with 28479 ligands (Figure 5). The estimated $\log K_{a,j}^{(\text{app})}$ of 28479 ligands largely distributed in a range between 3 and 5 (Figure 5(a)). The threshold $\log K_{a,j}^{(\text{app})}$ value for the resulting top-20 ligands (circles in Figure 5(b)) is 5.54 (dashed vertical line in Figure 5(b)). The docking times $n_j(T)$ among ligands are diverse in rUCB, so some estimated $\log K_{a,j}^{(\text{app})}$ are accurate while the others are not. To further confirm the effect of rUCB, we dock each top-20 ligand to extra 100 conformations of c-Myc, and the re-estimated $\log K_{a,j}^{(\text{app})}$ were shown in Figure 5(b) as error-bars. Although most $\log K_{a,j}^{(\text{app})}$ changes obviously in re-estimation, only two among top-20 ligands drop below the threshold value. The re-estimated $K_d = 1/K_{a,j}^{(\text{app})}$ of top-20 ligands lies between 0.7 μM and 6 μM, a half of which have binding affinity stronger than 2 μM, supporting the high effectiveness of rUCB.

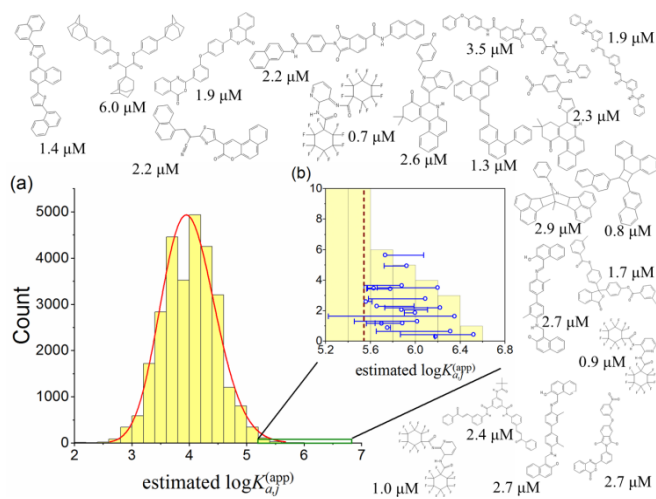

**Figure 5.** Application of rUCB on a library with $N = 28479$ ligands. $m = 20$, $T/N = 2$. (a) Histogram of estimated $\log K_{a,j}^{(\mathrm{app})}$ from rUCB running. (b) The estimated $\log K_{a,j}^{(\mathrm{app})}$ of the top-20 ligands being picked by rUCB (open circles), where the error bars indicate the re-estimated values from extra docking to 100 conformations for each picked ligands. Threshold value for top-20 ligands was shown as a dashed vertical line. Chemical structures of the top-20 ligands and their re-estimated $K_\mathrm{d} = 1/K_{a,j}^{(\mathrm{app})}$ are also shown.

In brief, we have developed rUCB for virtual screening upon IDPs. Compared with the well-developed drug design pipelines that target ordered proteins, the rational drug design for IDPs remain in its infancy largely because the well-developed structure-centered techniques can hardly be applied directly to IDPs. With rUCB, we demonstrated that the structure-based docking can be reorganized by reinforcement-learning algorithm to accomplish virtual screening upon conformation ensemble of IDPs. Best of all, the required computation cost to achieve excellent accuracy is comparable to that for ordered proteins. This great supports the optimism of rational drug design for IDPs. In addition, the idea of reinforcement learning may be extended as the other way round to refine structure-based docking with conformation ensemble of ligand.

## Acknowledgements

This work was supported by the National Natural Science Foundation of China (grant 21633001). Part of the analysis was performed on the High Performance Computing Platform of the Center for Life Science (Peking University). We thank Prof. Luhua Lai, and Qiaojing Huang for helpful discussions.

**Keywords:** intrinsically disordered proteins • molecular recognition • molecular docking • virtual screening

# Supporting Information

## Reinforcement Learning to Boost Molecular Docking upon Protein Conformational Ensemble

Bin Chong[1], Yingguang Yang[2], Zi-Le Wang[3], Hang Xing[1], and Zhirong Liu[1,*]

[1] College of Chemistry and Molecular Engineering, and Beijing National Laboratory for Molecular Sciences (BNLMS)
Peking University
Beijing 100871 (China)
[2] School of Cyberscience
University of Science and Technology of China
Hefei 230026 (China)
[3] State Key Laboratory of Low Dimensional Quantum Physics, Department of Physics
Tsinghua University
Beijing 100084 (China)

Correspondence and requests for materials should be addressed to Z.L.(e-mail: LiuZhiRong@pku.edu.cn).

## Calculation Procedures

**Binding site prediction.** We used the code CAVITY (Y. X. Yuan, J. F. Pei, L. H. Lai, *Curr. Pharm. Design* **2013**, *19*, 2326-2333) to determine the potential binding sites of each conformation for molecular docking. In case of multiple sites for one conformation, the pocket with the maximal potential binding affinity was chosen for docking.

**Molecular docking.** We used the open-source program Vina (O. Trott, A. J. Olson, *J. Comput. Chem.* **2010**, *31*, 455-461) for molecular docking. Default parameters were chosen, e.g., the grid size is 22×22×22, while the energy_range, exhaustiveness and num_modes are set to be 3, 8 and 8, respectively. The grid center is set to be the geometric center of the conformational cavities obtained from CAVITY. The docking score given by Vina is the binding free energy $\Delta G_{i,j}$ (in a unit of kcal/mol) between the conformation $i$ and the ligand $j$. Open Babel (N. M. O'Boyle, M. Banck, C. A. James, C. Morley, T. Vandermeersch, G. R. Hutchison, *J. Cheminformatics* **2011**, *3*, 33.) was used to preprocess the file format of proteins and ligands.

**Conformational ensemble of IDPs.** The conformational ensemble of the oncoprotein c-Myc was borrowed from a previous study of large-scale MD simulations (F. Jin, C. Yu, L. H. Lai, Z. R. Liu, *PLoS Comput. Biol.* **2013**, *9*, e1003249), giving 16716 conformations. More specifically, Hammoudeh *et al.* (D. I. Hammoudeh, A. V. Follis, E. V. Prochownik, S. J. Metallo, *J. Am. Chem. Soc.* **2009**, *131*, 7390-7401.) measured the chemical shifts and several NOE signal of c-Myc$_{370-409}$ with and without small molecules by NMR, and predicted the average dihedral angles of the main chain; Jin *et al.* utilized these angles to construct and optimized the single apo and holo structures which were further used as initial conformation in MD simulations; Jin *et al.* then conducted large-scale (with a total simulation time of 34.5 μs) replica-exchange MD using AMBER software and AMBER99SB force field, where the ionic strength of the system is set to 0.2 M.

**Ligand library.** The positive and negative control ligands (10074-A4 and AJ292) as well as 6 ligand molecules (YC-1101, YC-1201~YC-1205) obtained from virtual screening by Yu *et al.* (C. Yu, X. Niu, F. Jin, Z. Liu, C. Jin, L. Lai, *Sci. Rep.* **2016**, *6*, 22298) were selected in priority. Additional 275 ligands were selected from the library of SPECS, a worldwide provider of compound management services besides being a main supplier of screening compounds in drug discovery. Specifically, small compound set with amount of more than 50 mg, which contains ~140,000 compounds, were selected from the SPECS libraries. Based on the calculated fingerprint of each compound, Leader-Follower clustering and K-Mean clustering were conducted in sequence. Finally, the 275 target compounds were obtained. In total, 283 ligands were used in our study.

**Real dataset-I.** Each of the 16716 conformations from the ensemble of c-Myc was docked with each of the 283 ligands above, and $K_{a,j}^{(\text{app})}$ for each ligand is calculated with Eq.(2) of the main text. The distribution properties were analysed detailedly before (B. Chong, Y. G. Yang, C. G. Zhou, Q. J. Huang, and Z. R. Liu, preprint).

**Synthetic datasets.** The ensemble distribution of $\Delta G_{i,j}$ for each ligand $j$ is approximately described by a Gaussian distribution with a mean $\mu_j$ and a variance $\sigma_j^2$, i.e., $p(\Delta G_{i,j}) = \mathcal{N}(\Delta G_{i,j}|\mu_j, \sigma_j^2)$. The distributions of $\mu_j$ and $\sigma_j$ for ligands of the synthetic datasets are also assumed to be Gaussian, whose parameters were inferred from the real dataset-I with 283 ligands to give $p(\mu_j) = \mathcal{N}(\mu_j|-5.1, 0.65^2)$ and $p(\sigma_j) = \mathcal{N}(\sigma_j|0.44, 0.08^2)$ (in units of kcal/mol). The apparent association constant is calculated by $\ln K_{a,j}^{(\text{app})} = -\frac{1}{RT}\left(\mu_j - \frac{\sigma_j^2}{2RT}\right)$ according to the reference (B. Chong, Y. G. Yang, C. G. Zhou, Z. R. Liu, preprint).

# SUPPORTING INFORMATION

**Real dataset-II.** To provide a validation set, we selected additional 828 ligands randomly (which are different from those in real dataset-I) from the library of SPECS and docked them with all 16716 conformations from the ensemble of c-Myc. The rusults were used as the dataset to validate the effects of the rUCB algorithm.

**Large ligand library.** A large library with 28479 ligands (including dataset-I and dataset-II) from SPECS was constructed. The rUCB algorithm was used to screen this large library to target c-Myc.

## Details of the rUCB algorithm

The aim of the rUCB algorithm is to predict (pick out) top $m$ ligands with the largest $K_{a,j}^{(app)}$ from a library with $N$ ligands, by scheduling docking process with reinforcement learning approach, within a limited number of docking times $T$. Unless otherwise stated, $T = 2N$ is adopted in this study. The rUCB algorithm is composed of the following steps:

**Initialization:** Dock each ligand once, with IDP conformation randomly selected from the ensemble. The total number of docking times in this stage is $N$.

**Loop ($t = N + 1, T$):** Choose one ligand to dock once (with IDP conformation randomly selected from the ensemble) by

— Pre-choose $m$ ligands that maximize an indexed function as

$$\arg\max_j \left[ \log K_{a,j}^{(app)}(t) + c\sigma_{\log K,j}(t)\sqrt{\frac{1}{n_j(t)}} \right]. \tag{S1}$$

We adopt $Q_j = \log K_{a,j}^{(app)}$ with a logarithm operation to avoid the improper influence of the fact that $K_{a,j}^{(app)}$ spans a few orders of magnitude. $n_j(t)$ is the number of times that the ligand-$j$ has been docked prior to $t$. $\log K_{a,j}^{(app)}(t)$ is the estimate of $\log K_{a,j}^{(app)}$ based on the prior docking results, while $\sigma_{\log K,j}(t)$ is the estimated standard deviation of $\log K_{a,i,j}$. For $n_j(t) = 1$, a straightforward determination of $\sigma_{\log K,j}(t)$ is not available, then a default value of 0.35 inferred from the real docking dataset with 283 ligand is used. For in $t$ a range of $[N + 1:T]$ with $T = 2N$ and large $N$, the range of variation of the $\ln t$ term in usual UCB algorithm is very small, so we discard the $\ln t$ term. An advantage of this simplification is that we need not update the index function for all ligands at each step $t$, i.e., we need only update the index value of the ligand that was docked in the latest round and its ranking.

— Among $m$ pre-chosen ligands, choose the one that minimize another indexed function as

$$\arg\min_j \left[ \log K_{a,j}^{(app)}(t) - c\sigma_{\log K,j}(t)\sqrt{\frac{1}{n_j(t)}} \right], \tag{S2}$$

and dock it once (with IDP conformation randomly selected from the ensemble).

**Final prediction:** Pick $m$ ligands that maximize $\log K_{a,j}^{(app)}(T)$ as the predicted top ligands from virtual screening.

## Results and Discussion

### An Oversimplified Analysis on the docking times

By ignoring the difference of $\sigma_{\log K,j}$ for ligands, the indexed function in selecting ligands in Eq. (S1) becomes

$$\log K_{a,j}^{(app)}(t) + c'\sqrt{\frac{1}{n_j(t)}}. \tag{S3}$$

When the total docking number is large, the indexed function reaches a constant (denoted as $\log K_a^{(0)}$), so it yields

$$\log K_{a,j}^{(app)} + c'\sqrt{\frac{1}{n_j}} = \log K_a^{(0)}. \tag{S4}$$

So

$$n_j = \frac{c'^2}{\left(\log K_{a,j}^{(app)} - \log K_a^{(0)}\right)^2}. \tag{S5}$$

It diverges when $\log K_{a,j}^{(app)} = \log K_a^{(0)}$. In rUCB algorithm, $\log K_{a,j}^{(app)}$ is estimated from the prior docking results, which introducing fluctuation. So we add an extra term to reflect such an effect:

$$n_j = \frac{c'^2}{\left(\log K_{a,j}^{(app)} - \log K_a^{(0)}\right)^2 + \gamma^2}, \tag{S6}$$

which is an Cauchy-Lorentz function.

Analyses on Eq. (S2) gives similar result.

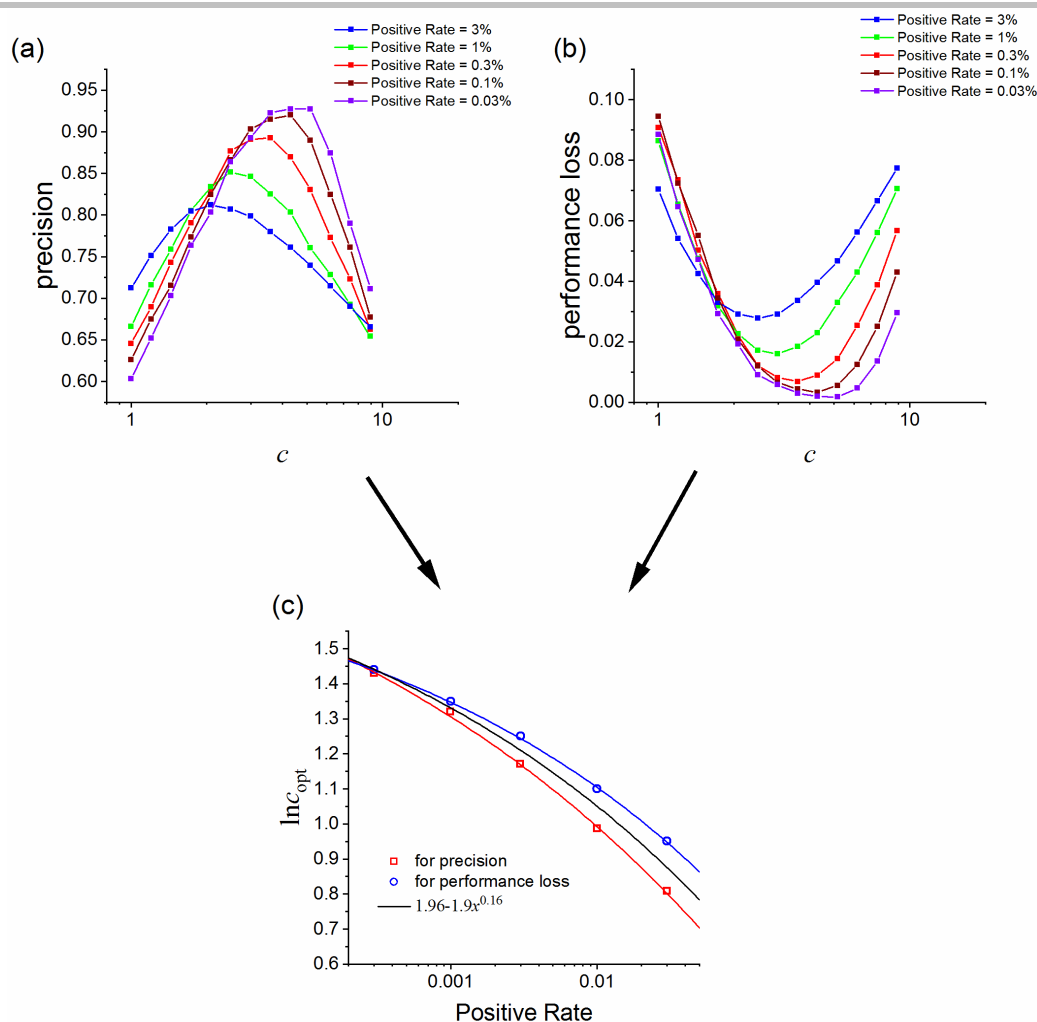

**Figure S1.** Influence of the learning parameter $c$ for the reversible UCB algorithm. Synthetic dataset with $N = 10^4$ is used. (a) The precision as a function of $c$. (b) The performance loss as a function of $c$. (c) The optimal $c$ as a function of the positive rate, where red squares were derived from (a) to achieve the maximal precision, and green circles were derived from (b) to achieve the minimal performance loss, while the black lines represents a compromise as $c = e^{1.96-1.9x^{0.16}}$ where $x$ is the positive rate defined as $m/N$. The total docking number in each run is two times the ligand number, i.e., $T = 2N$. Each data point in (a) and (b) was averaged from 200 runs of simulation with different random seeds.

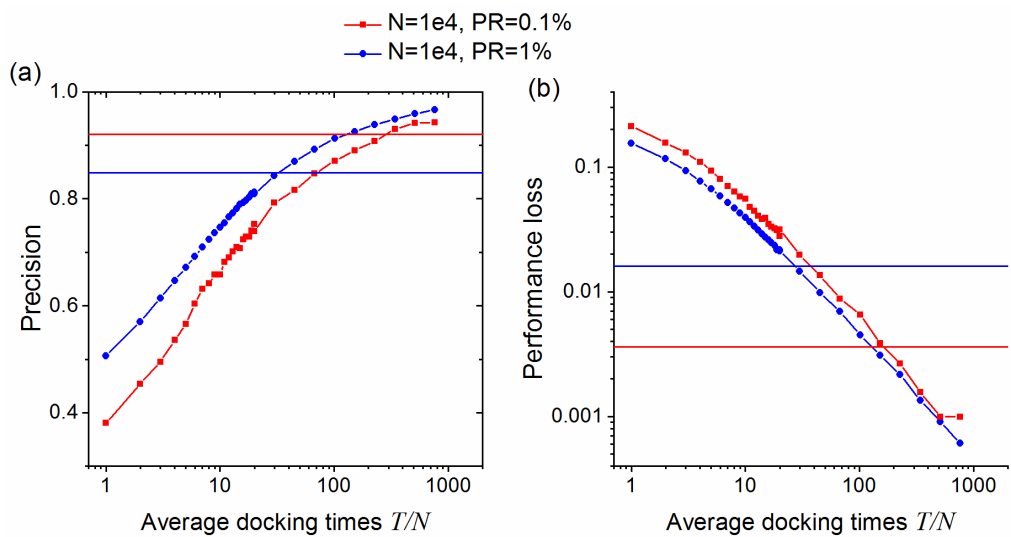

**Figure S2.** Effects of the uniform algorithm (curves with symbols) as functions of the average docking times per ligand, $T/N$. Synthetic dataset with $N = 10^4$ is used. The positive rate is 0.1% (curves with squares) or 1% (curves with circles). The rUCB results with $T/N = 2$ were shown as horizontal lines for comparison. Each data point was averaged from 200 runs of simulation.

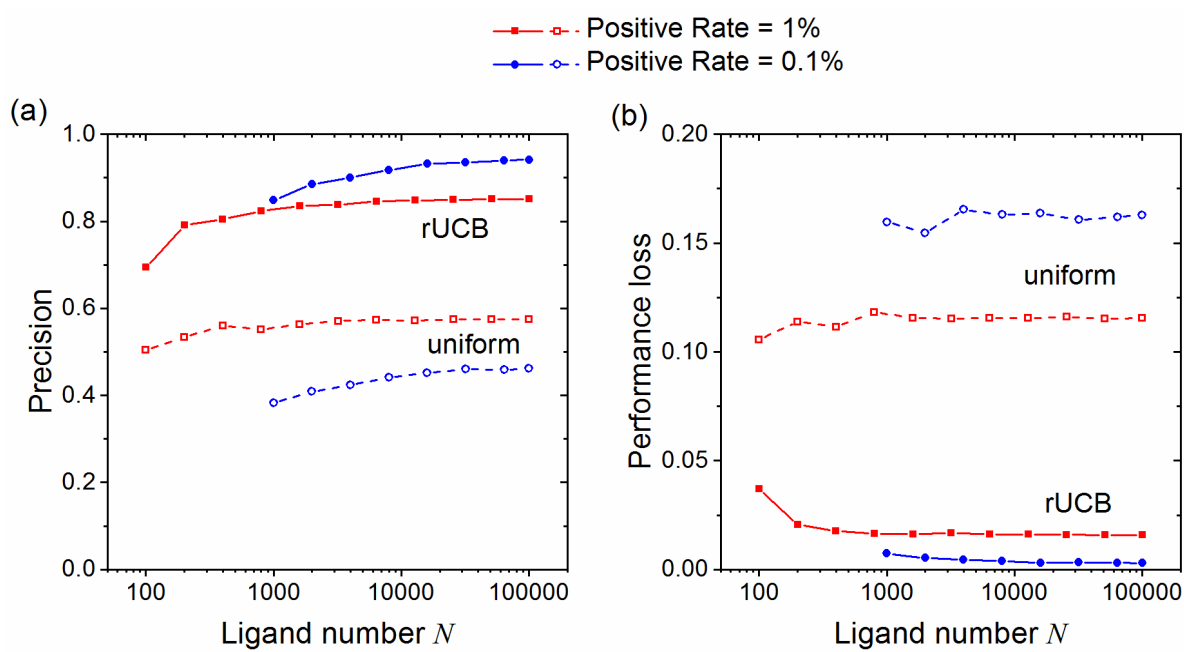

**Figure S3.** Effects of the rUCB algorithm (filled symbols) in comparison with those of the uniform algorithm (opened symbols) as functions of the ligand number $N$. Synthetic datasets with a positive rate of 1% (squares) and 0.1% (circles) were used. The average docking times per ligand is 2, i.e., $T/N = 2$. Each data point was averaged from 500 runs of simulation.

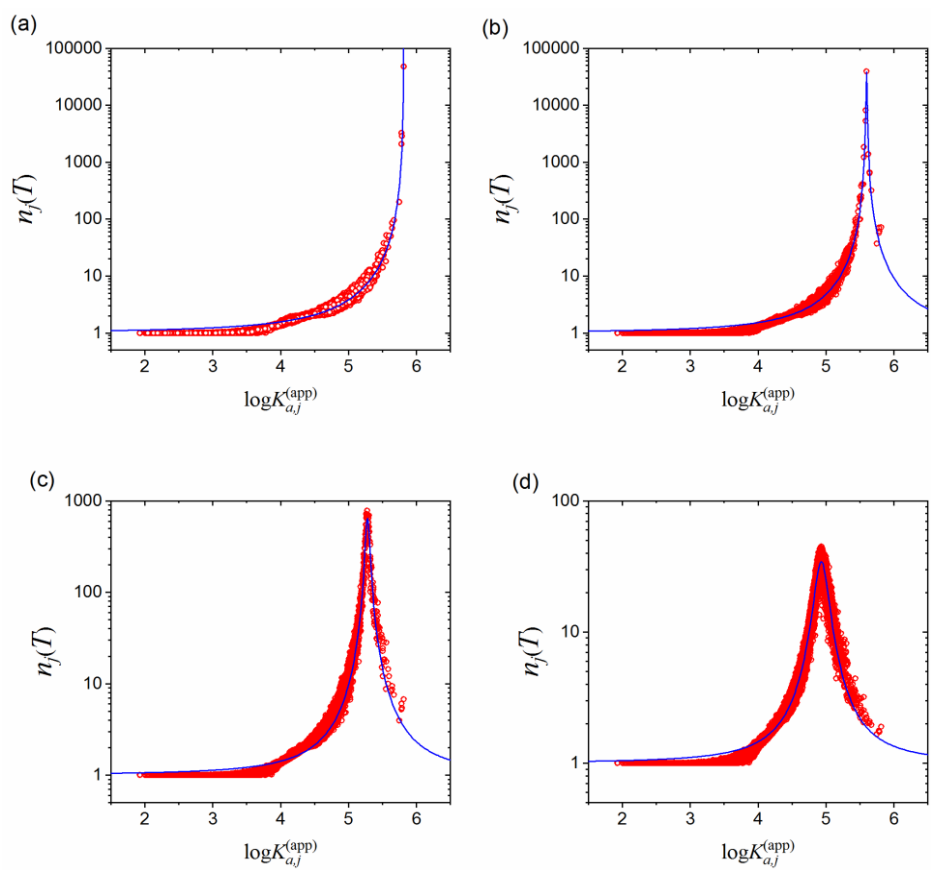

**Figure S4.** Distribution of docking times under rUCB for synthetic dataset with $N = 10^5$ and (a-d) $m = 1, 10, 100, 1000$, respectively, to give a positive rate of 0.001%, 0.01%, 0.1%, 1%. $T/N = 2$. Results from 400 simulations were averaged for each panel, where the blue line is the fitted Cauchy-Lorentz function.

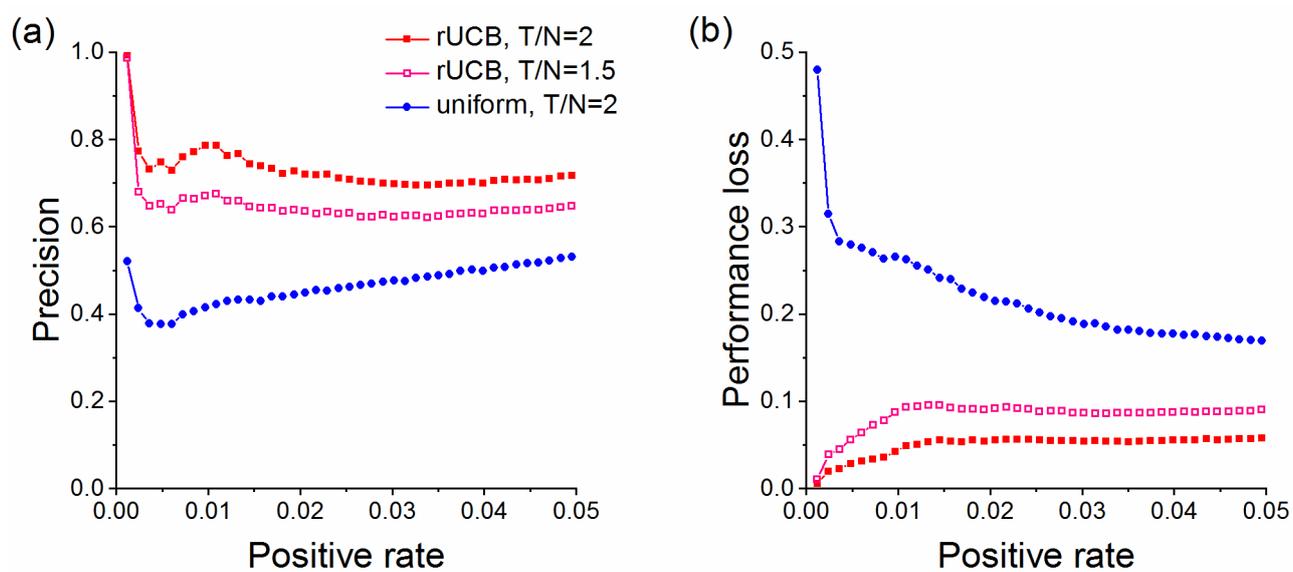

**Figure S5.** Validation of rUCB on the real dataset-II with 828 ligands. For each data point, 2000 runs of simulation were conducted to get average.